\documentclass[conference]{IEEEtran}
\IEEEoverridecommandlockouts
\usepackage[utf8]{inputenc}
\usepackage[T1]{fontenc}
\usepackage[english]{babel}
\usepackage{cite}
\usepackage{url}
\usepackage{xurl}
\usepackage{booktabs}
\usepackage{array}
\usepackage{tabularx}
\usepackage{multirow}
\usepackage{enumitem}
\usepackage{balance}
\usepackage[hidelinks]{hyperref}
\usepackage{microtype}
\hypersetup{pdftitle={The Integrator Advantage: Controlled Agentic AI for Small and Medium-Sized Companies}, pdfauthor={Christopher Koch and Joshua A. Wellbrock, Independent Researchers}, pdfsubject={Agentic AI integration for SMCs}}
\setlist[itemize]{leftmargin=*, topsep=2pt, itemsep=1pt}
\setlist[enumerate]{leftmargin=*, topsep=2pt, itemsep=1pt}
\newcolumntype{Y}{>{\raggedright\arraybackslash}X}
\newcolumntype{P}[1]{>{\raggedright\arraybackslash}p{#1}}
\title{The Integrator Advantage: Controlled Agentic AI for Small and Medium-Sized Companies\\
\large A Human-Centered Framework for Automating Simple and Medium-Complexity Processes}

\author{\IEEEauthorblockN{Christopher Koch}\\
\IEEEauthorblockA{Independent Researcher}\
\and\
\IEEEauthorblockN{Joshua A. Wellbrock}\\
\IEEEauthorblockA{Independent Researcher}}

\begin{document}
\maketitle

\begin{abstract}
Agentic AI marks a new phase of enterprise automation. Unlike traditional automation or conversational AI, agentic systems can interpret goals, plan multi step tasks, access tools, interact with enterprise systems, and execute workflows with varying degrees of autonomy. For small and medium sized companies, this creates potential to reduce administrative burden, accelerate routine processes, and improve the use of organizational knowledge. This paper argues that the near term value of Agentic AI does not lie in full autonomy or workforce reduction, but in controlled partial autonomy for simple and medium complexity business processes. It proposes an integration framework covering use case suitability, autonomy levels, technical integration, governance, security, employee enablement, and measurable impact. The paper concludes that Agentic AI can become a productivity lever when implemented as a human centered capability with responsibility and accountability retained by people.
\end{abstract}


\section{Introduction}
Small and medium-sized companies (SMCs) are under simultaneous pressure from labour shortages, administrative overload, fragmented information systems, regulatory complexity, cybersecurity risk, and rising expectations for service speed and quality. Many firms already run enterprise resource planning (ERP), customer relationship management (CRM), document management, ticketing, e-mail, and collaboration platforms. However, much of the real work still happens between these systems: employees search for information, copy data, interpret documents, compare records, prepare replies, create summaries, trigger approvals, and document outcomes. These connective tasks are often too variable for traditional robotic process automation (RPA) and too operational for a simple chatbot. They are the natural domain of agentic AI.

Recent evidence shows that AI adoption is broadening, but deep process integration is still immature. Bitkom reports that 36\% of German companies used AI in 2025, almost twice the previous year's share, while another 47\% planned or discussed adoption; the same study highlights legal uncertainty, missing technical know-how, lack of personnel resources, data protection, and acceptance as key barriers \cite{bitkom2025}. KfW Research finds that 20\% of German SMCs used AI in the period 2022 to 2024, compared with only 4\% in 2016 to 2018; among firms with more than 50 employees, the share reached 36\% \cite{kfw2026ki}. The OECD reports that 31\% of SMCs in seven advanced economies used generative AI in late 2024, with Germany at 39\%; among adopting SMCs, 65\% reported improved employee performance, while most did not report a reduction in staff needs \cite{oecd2025sme}.

The central problem is therefore no longer whether AI is relevant. The problem is how it becomes a reliable operating capability. McKinsey's 2025 global survey finds that almost all respondents report AI use, but most organizations remain in experimentation or piloting; 62\% are at least experimenting with agents, and workflow redesign is a key success factor among high performers \cite{mckinsey2025}. MIT Sloan Management Review and BCG likewise argue that agentic AI requires reimagining workflows, roles, governance, learning, and investment strategies because the technology combines properties of tools and of organizational actors \cite{bcg2025}. For SMCs, this points to a clear position: agentic AI should be introduced as controlled process capability, not as a vague autonomy project.

This paper uses the term \emph{integrator advantage} to describe the scarce capability that employers need when they search for people who can make agentic AI useful in practice. The integrator is not merely a prompt writer, data scientist, software engineer, or change manager. The integrator bridges process understanding, system interfaces, identity and access management, AI evaluation, governance, and workforce enablement. In agentic AI, the strategic bottleneck is often not access to a model but the ability to embed a model safely into work.

\subsection{Research Question and Contribution}
This position paper addresses the following question: \emph{How can SMCs integrate agentic AI into simple and medium-complexity business processes so that productivity, quality, and responsiveness increase while human responsibility, employability, and trust remain intact?}

The paper contributes five elements. First, it positions agentic AI as a controlled execution layer between people, knowledge, and enterprise systems. Second, it defines a use-case radar for simple, medium-complexity, and unsuitable processes. Third, it proposes the Agentic AI Integration Framework for SMCs (AIF-SMC). Fourth, it develops the Human Responsibility Core as a principle for responsibility allocation. Fifth, it offers an implementation roadmap, governance baseline, upskilling model, and KPI set that can guide first integrations without turning the paper into vendor marketing.

\section{Evidence Base and Analytical Lenses}
The paper is based on a structured synthesis of market studies, research on SMCs, workforce reports, agentic AI publications, business process management literature, and security and governance frameworks. Table \ref{tab:lenses} summarizes the ten analytical lenses used to converge on the position.

Methodologically, this paper is a conceptual positioning and framework-synthesis contribution. It does not report a new empirical field study, and the proposed framework should therefore be treated as an analytically grounded artefact that requires validation through case studies, expert interviews, or design science evaluation. The evidence cited supports increasing adoption, scaling challenges, and task transformation; the paper's anti-layoff stance is a strategic and normative positioning rather than an empirical guarantee that any deployment will preserve headcount.

\begin{table}[!t]
\centering
\caption{Ten analytical lenses for the position}
\label{tab:lenses}
\scriptsize
\begin{tabularx}{\columnwidth}{P{0.07\columnwidth} P{0.26\columnwidth} Y}
\toprule
No. & Lens & Synthesis result \\
\midrule
1 & Adoption in SMCs & AI use is rising, but process maturity and strategy lag. \\
2 & Agentic scaling & Agents remain early; workflow redesign separates pilots from value. \\
3 & Process fit & Best candidates are frequent, semi-structured, digital, and verifiable. \\
4 & Integration & Value depends on connecting tools, data, permissions, and approvals. \\
5 & Autonomy & Levels 2 to 4 are realistic; full autonomy is not a starting point. \\
6 & Governance & Owners, logs, access rights, monitoring, and kill switches are mandatory. \\
7 & Security & Agents need explicit identities and least-privilege access. \\
8 & Regulation & AI literacy and risk management are already operational obligations. \\
9 & Workforce & Tasks are recomposed; judgement and relationship work gain importance. \\
10 & Change & Upskilling and employee participation are part of the operating model. \\
\bottomrule
\end{tabularx}
\end{table}

The evidence base suggests a balanced position. Agentic AI is more capable than isolated generative AI assistance because it can plan and act across workflow steps. However, the same capability increases risk: agents can misuse tools, inherit excessive permissions, confuse data contexts, and produce plausible but incorrect outputs. Industrial evidence points to a capability-deployment verification gap. Apostolou et al. report that companies may demonstrate advanced experimental agentic capabilities but fail to deploy them into production because output verification mechanisms are absent; human-in-the-loop remains the only trusted verification mechanism in many settings \cite{apostolou2026}. This is precisely why integration, governance, and workforce design must be treated as first-class research and implementation topics.

\section{Agentic AI as an Integration Problem}
\subsection{Definition}
This paper defines agentic AI as an AI-based system that can interpret a goal, decompose it into intermediate steps, access relevant knowledge, call tools or systems, monitor intermediate outcomes, and return or execute a result within defined boundaries. The definition is intentionally operational. It excludes simple chat interfaces that only answer questions and includes systems that combine reasoning, tool use, memory or retrieval, workflow state, and escalation.

Agentic AI therefore differs from related automation patterns. RPA executes pre-defined steps against relatively stable interfaces. A chatbot answers user questions in a conversational interface. A copilot assists a human inside a task. An agentic workflow can combine all three: it can understand a request, retrieve evidence, compare it with system records, draft a response, create a ticket, request approval, and document the outcome. This makes agentic AI attractive for SMCs, where many processes are neither fully structured nor entirely creative.

\subsection{Why Integrators Matter}
The integrator advantage begins with the observation that agentic AI fails when it is treated either as a standalone model or as an uncontrolled digital employee. Real value requires integration across five layers: process, data, systems, governance, and people. An integrator translates an operational pain point into a bounded agent workflow; identifies which systems must be read or written; defines which data are authoritative; designs human approvals and escalation rules; implements test and monitoring logic; and prepares employees to supervise, challenge, and improve the agent.

This role is especially relevant for employers. Hiring or developing integrators means acquiring a capability to make AI operational, not merely decorative. A firm can buy a model or license a platform, but it cannot outsource the tacit knowledge of how invoices, customer exceptions, supplier negotiations, change requests, machine logs, quality deviations, and onboarding tasks actually work. The value is in combining domain reality with technical and organizational controls.

\begin{table}[!t]
\centering
\caption{The integrator competence stack}
\label{tab:integrator}
\scriptsize
\begin{tabularx}{\columnwidth}{P{0.31\columnwidth} Y}
\toprule
Competence & Practical signal in an agentic AI project \\
\midrule
Process literacy & Can map the real process, exceptions, handovers, decision rules, and pain points. \\
System integration & Understands ERP, CRM, DMS, e-mail, APIs, RPA fallbacks, data formats, and tool boundaries. \\
Agent design & Can define goals, tool access, prompts, retrieval, memory, state, tests, and escalation logic. \\
Security and governance & Applies least privilege, identity, audit trails, risk classification, monitoring, and incident handling. \\
Workforce enablement & Trains users, designs human-in-the-loop work, reduces fear, and creates feedback loops. \\
Business evaluation & Measures throughput, quality, override rate, cost per case, employee relief, and adoption. \\
\bottomrule
\end{tabularx}
\end{table}

\section{Current Studies and Implications}
\subsection{Small and Medium-Sized Companies as the Adoption Context}
OECD evidence is particularly relevant because it focuses on SMCs rather than only large enterprises. The OECD survey shows that generative AI has lowered the entry barrier for smaller firms, while still requiring structured support to close digital and skills gaps \cite{oecd2025sme}. For SMCs, this means agentic AI should not start as a large transformation program. It should start as a disciplined process intervention with clear human ownership.

Bitkom, KfW, and bidt point in the same direction. Adoption is growing, but strategies, competencies, and implementation depth lag behind \cite{bitkom2025}, \cite{kfw2026ki}, \cite{bidt2025}. KfW's Digitalization Report stresses the importance of digital strategy, digital pre-education, IT skills, lifelong learning, data protection, and data security for digitalization in SMCs \cite{kfw2025digital}. The implication is straightforward: agentic AI will not compensate for an absent digital foundation. It can accelerate a company that has enough process clarity and data access to govern it.

\subsection{Enterprise Scaling and Agentic AI}
McKinsey's 2025 survey shows broad AI adoption but limited enterprise-scale value capture; nearly two-thirds of respondents say their organizations have not begun scaling AI across the enterprise, and high performers redesign workflows rather than simply placing AI on top of existing work \cite{mckinsey2025}. Bain argues that agentic AI is a structural shift in enterprise technology because agents can reason, coordinate, and execute workflows, but capturing value requires modernizing architecture and scaling across core workflows \cite{bain2025foundation}. MIT SMR and BCG report that agentic AI is emerging rapidly and that firms must redesign workflows, roles, governance, and learning to manage systems that blur the boundary between tool and teammate \cite{bcg2025}.

Academic work on agentic business process management supports this view. Vu et al. describe agentic BPM as an area where generative AI-based agents can support processes, but they emphasize governance, human collaboration, transparency, and job displacement concerns \cite{vu2025}. Azarijafari et al. propose a goal-oriented approach to business process development in which agents contribute to business goals rather than merely executing fixed tasks \cite{azarijafari2025}. For SMCs, these contributions suggest that process architecture and governance should precede tool enthusiasm.

\subsection{Workforce Transformation}
The workforce literature argues against a simplistic substitution narrative. The OECD finds that SMCs using generative AI most often report improved employee performance; the effect on headcount is limited in the near term \cite{oecd2025sme}. The ILO's 2025 update examines occupational exposure at task level and frames generative AI impact as transformation of job content rather than automatic elimination of jobs \cite{ilo2025}. The World Economic Forum expects major labour-market churn by 2030, with 22\% of jobs structurally affected, 170 million roles created, and 92 million displaced globally \cite{wef2025}. PwC's 2026 AI Jobs Barometer reports that companies most exposed to AI have seen higher productivity growth and that the skills required in AI-exposed jobs are changing more than twice as fast as in less exposed jobs; judgement, leadership, empathy, and creativity become more important as AI absorbs routine work \cite{pwc2026}. Microsoft similarly argues that as agents take on execution, human agency can expand if organizations redesign work and learning systems accordingly \cite{microsoft2026}.

For SMCs, the conclusion is clear: agentic AI should be framed as capacity expansion and task recomposition, not as a layoff program. That framing is not only ethical but economically rational. Many SMCs face shortages of skilled people. Their constraint is not that they employ too many domain experts; it is that domain experts spend too much time on repetitive coordination, documentation, and searching.

\section{Agentic AI Integration Framework for SMCs}
The Agentic AI Integration Framework for SMCs consists of five mutually dependent components: use-case suitability, autonomy level, technical integration, governance, and employee enablement. The framework is designed for practical use by firms that do not have large AI departments but still need disciplined deployment.

\subsection{Component 1: Use-Case Suitability}
Agentic AI is best suited to processes that are frequent, semi-structured, digital, verifiable, and limited in downside risk. Simple use cases include e-mail classification, meeting and ticket summarization, document extraction, FAQ drafting, report condensation, and routing. Medium-complexity use cases combine several systems or judgement steps: invoice validation against purchase order and goods receipt, customer inquiry resolution using CRM and ERP data, offer preparation, supplier comparison, HR onboarding coordination, quality deviation pre-analysis, and service incident triage.

Unsuitable starting points include employment decisions, salary decisions, autonomous high-value payments, legal commitments, safety-critical production control, employee surveillance, and strategic decisions with high uncertainty. These may involve AI assistance in the future, but not early agentic automation.

\begin{table}[!t]
\centering
\caption{Use-case radar for agentic AI for SMCs}
\label{tab:usecases}
\scriptsize
\begin{tabularx}{\columnwidth}{P{0.20\columnwidth} P{0.28\columnwidth} Y}
\toprule
Domain & Simple entry point & Medium-complexity target \\
\midrule
Customer service & Classify requests and draft replies & Resolve standard cases using CRM, ERP, and policy data. \\
Finance & Extract invoice data & Validate invoice, purchase order, and receipt; prepare exception report. \\
Procurement & Collect supplier information & Compare offers, flag risks, prepare negotiation brief. \\
Sales & Summarize calls & Prepare offer draft from CRM, product data, and templates. \\
HR & Answer onboarding FAQs & Coordinate onboarding tasks, documents, appointments, and access requests. \\
IT & Route tickets & Analyze logs, suggest fix, prepare change or incident note. \\
Quality & Structure deviation text & Draft root-cause hypotheses and CAPA preparation. \\
Operations & Summarize shift notes & Create maintenance ticket from logs, history, and thresholds. \\
\bottomrule
\end{tabularx}
\end{table}

\subsection{Component 2: Autonomy Levels}
The autonomy ladder in Table \ref{tab:autonomy} prevents two common errors: under-ambition, where AI remains only a writing assistant, and over-ambition, where autonomy is introduced before controls exist. The recommended target range for SMCs is Level 2 to Level 4.

\begin{table}[!t]
\centering
\caption{Autonomy ladder for controlled deployment}
\label{tab:autonomy}
\scriptsize
\begin{tabularx}{\columnwidth}{P{0.08\columnwidth} P{0.30\columnwidth} Y}
\toprule
Level & AI role & Recommendation for SMCs \\
\midrule
0 & No AI automation & Baseline for measurement. \\
1 & Assistant that answers, summarizes, or drafts & Good entry point; low integration risk. \\
2 & Proposal engine that prepares decisions & Highly suitable for many administrative processes. \\
3 & Tool-using agent with human approval & Core target for medium-complexity workflows. \\
4 & Supervised standard-case automation with escalation & Suitable after stable metrics and governance. \\
5 & Full autonomy with minimal human intervention & Not a starting point for most SMC contexts. \\
\bottomrule
\end{tabularx}
\end{table}

\subsection{Component 3: Technical Integration}
A practical target architecture contains seven layers. First, a human process owner defines purpose, process boundaries, quality criteria, and escalation rules. Second, an agent orchestrator plans steps and calls tools. Third, system connectors provide scoped access to ERP, CRM, DMS, e-mail, ticketing, quality management, or production data. Fourth, a knowledge layer provides approved procedures, templates, policies, product information, contract rules, and FAQs. Fifth, a policy and permission layer controls data classes, roles, actions, and approval thresholds. Sixth, evaluation and monitoring track quality, cost, latency, hallucination classes, tool use, and overrides. Seventh, human-in-the-loop work ensures approval, exception handling, and accountability.

Three integration patterns are especially useful. A \emph{read-first agent} can search, classify, compare, and draft but cannot write back to systems. A \emph{prepare-and-approve agent} prepares records, e-mails, tickets, or booking proposals for human confirmation. A \emph{standard-case agent} executes narrowly defined standard cases autonomously but escalates every deviation. These patterns create a path from trust building to measured autonomy.

\subsection{Component 4: Governance}
Governance is not bureaucracy added after innovation. It is the condition that allows autonomy to leave the pilot. A minimal SMC governance set should include an agent register, business owner, technical owner, purpose statement, risk classification, data classification, permission model, audit trail, human-in-the-loop rules, escalation logic, quality monitoring, incident process, change management for prompts, models, tools, and retrieval sources, and a kill switch.

Regulation and standards already provide useful anchors. Article 4 of the EU AI Act requires providers and deployers of AI systems to ensure an adequate level of AI literacy for staff and other persons dealing with operation and use \cite{euaiart4}. The EU also introduced obligations for general-purpose AI model providers from August 2025, including documentation and additional requirements for systemic-risk models \cite{eugpai2025}. NIST's Generative AI Profile extends the AI Risk Management Framework to generative AI risks \cite{nist2024}. ISO/IEC 42001 specifies requirements for establishing and maintaining an AI management system \cite{iso42001}. OWASP's Top 10 for Agentic Applications provides a security-oriented reference for autonomous, tool-using systems \cite{owasp2025}.

Agent identity is a particularly important topic. The OpenID Foundation highlights that AI agents raise new questions in authentication, authorization, delegated authority, and identity management \cite{openid2025}. The Cloud Security Alliance reports that 68\% of surveyed organizations cannot clearly distinguish human from AI-agent activity and that identity and access models have not kept pace with agent autonomy \cite{csa2026}. Therefore, a production agent should not operate through a shared service account or permanently under a human identity. It should have a distinct identity, explicit delegation, least-privilege permissions, auditable actions, and revocable credentials.

\subsection{Component 5: Employee Enablement}
Employee enablement is the part most often under-designed. A firm that automates without training creates fear, dependency, and hidden correction work. A firm that trains employees only in prompting misses the broader skill set. Agentic AI requires AI literacy, process understanding, verification skills, data judgement, escalation discipline, and an understanding of responsibility boundaries.

The recommended upskilling stack has four levels. Level 1 is AI literacy for all employees: capabilities, limits, data protection, hallucinations, source checking, and safe use. Level 2 is role-specific usage: for example invoice validation in finance, answer review in service, or deviation analysis in quality. Level 3 is agent supervision: test cases, feedback, monitoring, escalation, and output review. Level 4 is governance and operations: identity, permissions, audit, incident response, cost control, and model/tool change management.

\section{The Integrator Operating Model}
\subsection{From Tool Rollout to Work Redesign}
A recurring implementation failure is to treat agentic AI as a software rollout. In that view, the organization buys a tool, grants access, sends a short training link, and expects productivity to follow. Agentic AI requires a different operating model because the system can initiate tool calls, combine data, produce intermediate artefacts, and influence decisions. The unit of design is not the tool but the work system around it.

The integrator operating model contains four artefacts. The first is an \emph{agent charter}: a one-page description of the agent's purpose, users, data sources, allowed actions, forbidden actions, human owner, success metrics, and shutdown procedure. The second is a \emph{process map}: a representation of the current and target workflow, including exceptions and handovers. The third is a \emph{control plan}: permission model, test cases, approval thresholds, audit data, escalation rules, and incident handling. The fourth is a \emph{learning plan}: training, feedback, office hours, user support, and a cadence for improving prompts, tools, retrieval sources, and process rules.

These artefacts are deliberately lightweight. Their purpose is not to create enterprise bureaucracy. Their purpose is to make autonomy discussable. If a firm cannot describe what an agent is allowed to do, what it must never do, and how humans can detect and correct failure, the agent is not ready for production.

\begin{table}[!t]
\centering
\caption{Core artefacts of the integrator operating model}
\label{tab:artefacts}
\scriptsize
\begin{tabularx}{\columnwidth}{P{0.25\columnwidth} P{0.33\columnwidth} Y}
\toprule
Artefact & Minimum content & Why it matters \\
\midrule
Agent charter & Purpose, owner, user group, data scope, tool scope, no-go actions, KPIs & Converts an AI idea into an accountable process component. \\
Process map & Current work, target work, handovers, exceptions, decisions, approvals & Prevents automation of misunderstood work. \\
Control plan & Access, logs, thresholds, tests, escalation, kill switch, incident path & Allows safe increase of autonomy. \\
Learning plan & Training, feedback loops, user support, change cadence, role development & Turns adoption into capability building. \\
\bottomrule
\end{tabularx}
\end{table}

\subsection{RACI for Controlled Autonomy}
Agentic AI should not belong only to IT or only to the business. Ownership must be shared but not vague. Table \ref{tab:raci} proposes a minimal responsibility model. The business process owner is accountable for the process outcome. IT is accountable for stable and secure integration. Security and data protection define guardrails. HR owns learning and workforce transition. The integrator coordinates the translation across these domains.

\begin{table}[!t]
\centering
\caption{Minimal RACI logic for a process agent in an SMC}
\label{tab:raci}
\scriptsize
\begin{tabularx}{\columnwidth}{P{0.24\columnwidth} P{0.18\columnwidth} Y}
\toprule
Activity & Accountable role & Consulted roles \\
\midrule
Use-case selection & Business owner & Integrator, team lead, IT, HR. \\
Process redesign & Business owner & End users, integrator, quality or compliance. \\
System connection & IT owner & Integrator, security, vendor. \\
Permission model & Security or IAM owner & IT, business owner, data protection. \\
Training plan & HR or learning owner & Integrator, business owner, team leads. \\
Quality monitoring & Business owner & Integrator, AI quality reviewer, IT. \\
Incident response & IT/security owner & Business owner, legal, data protection. \\
Scale decision & Management & All owners, employee representatives where applicable. \\
\bottomrule
\end{tabularx}
\end{table}

\section{Use-Case Selection and Scoring}
\subsection{A Practical Suitability Score}
An SMC does not need a complex AI portfolio office to start. It needs a transparent selection score that prevents political or hype-driven choices. Each candidate process can be rated from 1 to 5 across nine dimensions: frequency, manual time, standardizability, data availability, system access, verifiability, risk, employee acceptance, and economic value. Risk should be inverted in the total score: lower risk increases suitability. A simple formula is sufficient:

\begin{equation}
S = F + T + St + D + A + V + (6-R) + E + B
\end{equation}

where $F$ is frequency, $T$ is manual time, $St$ is standardizability, $D$ is data availability, $A$ is system access, $V$ is verifiability, $R$ is risk, $E$ is employee acceptance, and $B$ is business value. The point of the score is not mathematical precision. It creates a shared conversation about why one process should be piloted before another.

\begin{table}[!t]
\centering
\caption{Interpretation of the Agentic AI Suitability Score}
\label{tab:score}
\scriptsize
\begin{tabularx}{\columnwidth}{P{0.18\columnwidth} Y}
\toprule
Score range & Recommended decision \\
\midrule
38--45 & Strong pilot candidate. Start with Level 2 or 3 and prepare scale path. \\
30--37 & Suitable with process clarification and governance. Use shadow mode. \\
22--29 & Use assistance only; improve data, process rules, or verification first. \\
Below 22 & Avoid agentic automation for now; revisit after digital foundation improves. \\
\bottomrule
\end{tabularx}
\end{table}

\subsection{Two Reference Blueprints}
The first reference blueprint is a customer request agent. The agent reads incoming e-mails, classifies intent, retrieves customer history and order status, checks policy rules, drafts a reply, and proposes next steps. It may create an internal ticket but cannot send external communication without approval during the first production stage. Escalation is mandatory for complaints, legal threats, price exceptions, special customers, personal data requests, and uncertainty above a defined threshold. The value is reduced response time and better preparation, not replacing service judgement.

The second reference blueprint is an invoice validation agent. The agent extracts invoice data, checks supplier master data, compares the invoice with purchase order and goods receipt, flags mismatches, drafts an exception note, and prepares a booking or approval proposal. It cannot release payment autonomously above a threshold and cannot change master data. The value is reduced manual checking and faster exception handling. Finance employees retain accountability for posting, payment, and supplier communication.

Both blueprints demonstrate why integrators are valuable. They require process logic, system connections, data classification, approval thresholds, exception categories, user training, monitoring, and communication. The model is not ``deploy an AI agent''. The model is ``design a controlled work system in which an AI agent is one component''.

\section{Risk Model and Control Patterns}
Agentic AI risk is not only output hallucination. Once an agent uses tools, risk expands into action, identity, permission, and process effects. Table \ref{tab:risks} summarizes key risks and control patterns. The table draws on agentic AI security discussions, NIST risk management, OWASP agentic guidance, and identity-management work \cite{nist2024}, \cite{owasp2025}, \cite{openid2025}.

\begin{table}[!t]
\centering
\caption{Risk model for controlled agentic AI}
\label{tab:risks}
\scriptsize
\begin{tabularx}{\columnwidth}{P{0.24\columnwidth} P{0.32\columnwidth} Y}
\toprule
Risk & Typical manifestation & Control pattern \\
\midrule
Hallucination & Plausible but unsupported answer or analysis & Retrieval with citations, test sets, sampling, human review. \\
Tool misuse & Agent calls wrong tool or performs wrong action & Tool allowlist, action schemas, approval thresholds. \\
Prompt injection & Malicious content changes agent behaviour & Input isolation, instruction hierarchy, content filtering, no direct tool execution from untrusted text. \\
Over-privilege & Agent has broader access than task requires & Distinct identity, least privilege, short-lived credentials. \\
Data leakage & Sensitive data enters external model or wrong context & Data classification, local routing rules, redaction, vendor review. \\
Non-determinism & Same case produces inconsistent output & Test suites, confidence thresholds, deterministic rules for critical steps. \\
Over-trust & Users stop checking outputs & Training, sampling audits, visible evidence, override culture. \\
Botsitting & Users spend hidden time correcting agents & Measure correction time, improve process, reduce unnecessary autonomy. \\
Agent sprawl & Many unmanaged agents emerge & Agent register, lifecycle management, owner review. \\
\bottomrule
\end{tabularx}
\end{table}

A useful rule is that autonomy should increase only when controls become stronger. If a use case moves from drafting to tool execution, identity and permission controls must improve. If it moves from approval-based execution to standard-case automation, monitoring, sampling, incident handling, and rollback must improve. Controlled autonomy is therefore not a single state. It is a maturity path.

\section{Legacy System Integration Patterns}
System landscapes in SMCs often contain a mixture of modern SaaS systems, older ERP modules, local files, spreadsheets, supplier portals, e-mail-based approvals, and informal knowledge. The integration strategy should be pragmatic but not careless. API-based integration is preferable when available because it is more stable and auditable than UI automation. RPA can be useful where APIs do not exist, but it should be treated as a brittle adapter, not as invisible infrastructure. Retrieval-augmented generation can make policies and documents usable, but only if sources are curated and stale information is removed.

\begin{table}[!t]
\centering
\caption{Integration patterns for heterogeneous SMC systems}
\label{tab:patterns}
\scriptsize
\begin{tabularx}{\columnwidth}{P{0.25\columnwidth} P{0.32\columnwidth} Y}
\toprule
Pattern & Use when & Design caution \\
\midrule
API connector & ERP, CRM, DMS, or ticketing system exposes stable endpoints & Use scoped tokens, schema validation, and audit logs. \\
Read-only retrieval & Policies, manuals, contracts, FAQs, and product data are document-heavy & Curate sources; remove outdated documents; show evidence. \\
Prepare-and-approve writeback & Agent creates drafts or structured records & Require human approval before external or financial action. \\
RPA adapter & No API exists and UI steps are stable enough & Monitor fragility; avoid safety-critical dependence. \\
Human handoff & Ambiguity, risk, or missing data exceeds threshold & Preserve context and evidence when escalating. \\
Event trigger & New e-mail, ticket, invoice, or sensor event starts workflow & Validate input source and isolate untrusted content. \\
\bottomrule
\end{tabularx}
\end{table}

The best integrators know when not to integrate. If a system has poor data quality, unclear ownership, or unstable interfaces, the first step may be data and process cleanup rather than agent deployment. This restraint is part of professional credibility. It avoids creating a clever front end over a fragile back end.

\section{Implementation Playbook}
\subsection{Strategic Principles}
An implementation in an SMC should start with a simple management statement: \emph{We automate repetitive tasks, not people.} This statement must be operationalized, not merely communicated. Early automation waves should target work with high repetition, low meaning, high administrative burden, or high error risk. Productivity gains should be reinvested first into quality, customer response, growth capacity, documentation robustness, employee relief, and training.

The second principle is bounded autonomy. Every agent needs an explicit purpose, data scope, tool scope, action scope, approval logic, and shutdown mechanism. The third principle is evidence-before-expansion. A pilot should only scale when baseline metrics, user feedback, error classes, security review, and governance are in place.

\subsection{A 90-Day Plan}
Table \ref{tab:90days} proposes a 90-day plan for a first controlled process agent. It is deliberately short enough to avoid strategy paralysis and structured enough to avoid unmanaged experimentation.

\begin{table}[!t]
\centering
\caption{Ninety-day plan for the first controlled process agent}
\label{tab:90days}
\scriptsize
\begin{tabularx}{\columnwidth}{P{0.18\columnwidth} P{0.32\columnwidth} Y}
\toprule
Phase & Output & Key decisions \\
\midrule
Days 1--15 & Mandate, no-go rules, owners, communication narrative & Define task automation, not people replacement; set autonomy boundaries. \\
Days 16--30 & Process inventory and shortlist & Score volume, risk, verifiability, data access, and acceptance. \\
Days 31--45 & Pilot design and data approval & Define test data, connectors, tool allowlist, privacy and security review. \\
Days 46--70 & Prototype and shadow mode & Agent runs next to humans; measure quality, errors, effort, and trust. \\
Days 71--85 & Human-in-the-loop production test & Standard cases require approval; escalations and kill switch are tested. \\
Days 86--90 & Go/no-go and scale decision & Scale only if quality, governance, and employee impact are acceptable. \\
\bottomrule
\end{tabularx}
\end{table}

Shadow mode is the most important discipline. The agent processes real or realistic cases next to the existing process, but without autonomous external effect. This reveals missing data, hallucination classes, fragile integrations, false escalations, prompt injection issues, and hidden employee correction work before autonomy increases. The practice directly addresses the verification gap observed in industrial adoption studies \cite{apostolou2026}.

\subsection{Decision Matrix}
Every use case should end in one of four decisions: assist, pilot, automate, or avoid. Table \ref{tab:decision} summarizes the logic.

\begin{table}[!t]
\centering
\caption{Decision matrix for agentic AI use cases}
\label{tab:decision}
\scriptsize
\begin{tabularx}{\columnwidth}{P{0.21\columnwidth} P{0.34\columnwidth} Y}
\toprule
Decision & Criteria & Example \\
\midrule
Assist & Low integration maturity, uncertain data, human judgement required & Summaries, drafts, research briefs. \\
Pilot & Frequent, painful, measurable, multi-system, approval possible & Service triage, offer preparation, invoice exception report. \\
Automate & Stable standard cases, digital data, bounded risk, audit and escalation established & Ticket routing, document classification, simple validation. \\
Avoid & High legal, financial, safety, employment, or surveillance risk & Dismissal, autonomous payment, safety control, employee monitoring. \\
\bottomrule
\end{tabularx}
\end{table}

\section{Employee Impact: Responsibility, Trust, and Upskilling}
\subsection{From Substitution to Task Recomposition}
Agentic AI changes work at the task level. A role usually contains routine execution, search, coordination, documentation, exception handling, relationship management, judgement, accountability, and learning. Agents can take over parts of search, drafting, classification, reconciliation, and standard execution. Humans should retain context, values, responsibility, exceptions, relationships, and learning. The productive question is therefore not ``Which people can be replaced?'' but ``Which scarce human time can be multiplied?''

This distinction matters for trust. Employees are more likely to engage when the first use cases remove low-value burden rather than threaten identity, status, or job security. An SMC should communicate that agentic AI is introduced to reduce repetitive load, increase quality, and make expertise more scalable. That message must be supported by practical rights: the right to see what the agent did, the right to override outputs, the right to escalate, and the right to receive training before productivity expectations change.

\subsection{Human Responsibility Core}
The Human Responsibility Core is the principle that agents may execute tasks, but accountability remains assigned to human roles and organizational units. It contains eight elements: goal-setting, contextual judgement, value judgement, approval, exception handling, relationship management, liability, and continuous improvement. An agent may prepare an invoice validation, but finance owns booking and payment. An agent may draft a customer reply, but service owns the relationship and the tone. An agent may structure a corrective action proposal, but quality management owns the decision.

Human-in-the-loop must not turn employees into liability buffers for poorly designed systems. If a human is expected to review an output, the organization must provide time, criteria, access to evidence, training, and authority to reject or correct it. Otherwise the human is formally responsible but practically powerless.

\subsection{New Roles in Human-Agent Teams}
Agentic AI creates new roles inside business departments, not only in IT. Table \ref{tab:roles} maps existing roles to human-agent responsibilities.

\begin{table}[!t]
\centering
\caption{New roles in human-agent teams}
\label{tab:roles}
\scriptsize
\begin{tabularx}{\columnwidth}{P{0.28\columnwidth} P{0.29\columnwidth} Y}
\toprule
Current role & Emerging role & Main responsibility \\
\midrule
Clerk or case handler & Process reviewer & Exceptions, quality, approval, and feedback. \\
Customer service agent & Customer resolution owner & Relationship, tone, goodwill, and complex cases. \\
Buyer & Supplier intelligence owner & Risk, comparison, negotiation preparation. \\
Controller & Business insight partner & Interpretation, scenarios, management dialogue. \\
HR specialist & Employee journey coordinator & Sensitive cases, culture, onboarding experience. \\
IT support & Agent operations owner & Monitoring, incidents, access, and tool reliability. \\
Quality manager & AI-supported quality reviewer & Evidence, root cause, CAPA, and compliance judgement. \\
Team lead & Human-agent team lead & Work redesign, coaching, prioritization, learning. \\
\bottomrule
\end{tabularx}
\end{table}

These roles create a positive employment narrative. The employee is not downgraded to a passive checker. The employee becomes a supervisor of quality, context, and improvement. This is also where employers can identify strong integrators: people who naturally connect operational detail with systems thinking, risk awareness, communication, and learning.

\subsection{Communication Contract and Anti-Patterns}
A human-centered position must be translated into explicit commitments. The recommended communication contract has five clauses. First, the first wave of agentic AI targets repetitive, administrative, and low-meaning tasks before it touches identity-forming expert work. Second, employees are involved in selecting use cases because they know where friction, rework, and unnecessary search effort occur. Third, employees receive training time before new productivity expectations are set. Fourth, agents are not introduced as hidden performance surveillance systems. Fifth, productivity gains are made visible and reinvested into service capacity, quality, growth, learning, or workload relief.

This contract is not a soft add-on. It directly affects adoption quality. Employees who fear replacement or surveillance may withhold process knowledge, avoid experimentation, or create shadow practices. Employees who experience control, transparency, and skill development are more likely to improve the agent and contribute exceptions that no initial designer could anticipate.

\begin{table}[!t]
\centering
\caption{Implementation anti-patterns and better alternatives}
\label{tab:antipatterns}
\scriptsize
\begin{tabularx}{\columnwidth}{P{0.28\columnwidth} P{0.31\columnwidth} Y}
\toprule
Anti-pattern & Why it fails & Better alternative \\
\midrule
Tool-first rollout & Solves no specific process pain and creates low-value experimentation & Start from high-friction workflows and define measurable outcomes. \\
Cost-cutting narrative & Creates fear and reduces willingness to share process knowledge & Frame as capacity, quality, and skill expansion. \\
Invisible autonomy & Users do not know what the agent did or why & Provide logs, evidence, explanations, and approval points. \\
Human as liability buffer & Employees carry responsibility without time or authority & Give review time, criteria, override rights, and escalation paths. \\
Prompt-only training & Ignores verification, data judgement, and process responsibility & Train AI literacy, role use, supervision, and governance. \\
Unmanaged agent sprawl & Creates security, cost, and accountability gaps & Maintain agent register, owner model, and lifecycle reviews. \\
\bottomrule
\end{tabularx}
\end{table}

\subsection{Career Paths Instead of Layoff Logic}
An employer in an SMC can use agentic AI to create credible career paths. Domain experts can become knowledge curators, process reviewers, or agent supervisors. Technically oriented employees can become integration owners or agent operations specialists. Team leaders can become human-agent team leads who manage throughput, learning, and exception quality. These paths matter because they transform adoption from threat into progression.

The career-path logic also supports retention. Employees with operational knowledge are difficult to replace. If agentic AI is used to remove their repetitive burden while increasing their responsibility for judgement, customer outcomes, and process improvement, the firm preserves institutional knowledge and makes it more scalable. If the same technology is used mainly to cut headcount, the firm may save short-term cost but lose the human context required to govern and improve the agents.

\section{KPIs and Evaluation}
A credible integration paper must define success beyond cost reduction. The KPI model should combine process, employee, governance, and economic metrics. Process metrics include cycle time, manual touch time, first-time-right rate, backlog, response time, rework, and escalation rate. Employee metrics include training coverage, perceived relief, trust in the agent, confidence in override rights, time spent on higher-value work, and hidden correction time. Governance metrics include registered agents, audit completeness, permission violations, incident count, override rate, model or prompt change frequency, and cost per case. Economic metrics include throughput increase, capacity created, service-level improvement, avoided overtime, quality cost reduction, and revenue enablement.

The most important KPI may be the override rate. A very high override rate indicates poor agent quality or unclear process rules. A very low override rate may signal either excellent performance or dangerous over-trust. It must be interpreted together with sampling audits and error severity. The second important KPI is hidden work: if employees spend large amounts of time correcting, feeding, or monitoring the agent, apparent productivity gains are misleading. The third is adoption quality: a process agent that is not trusted, understood, or used correctly will not scale.

\begin{table}[!t]
\centering
\caption{Balanced KPI set for controlled agentic AI}
\label{tab:kpis}
\scriptsize
\begin{tabularx}{\columnwidth}{P{0.24\columnwidth} P{0.32\columnwidth} Y}
\toprule
Category & KPI examples & Interpretation \\
\midrule
Process & Cycle time, manual touch time, first-time-right, backlog & Shows operational improvement. \\
Quality & Error severity, rework, source accuracy, evidence completeness & Prevents speed without reliability. \\
Employee & Training coverage, perceived relief, trust, hidden correction time & Tests whether work becomes better. \\
Governance & Agent register, audit completeness, override rate, incidents & Shows whether autonomy is controlled. \\
Economics & Cost per case, capacity created, service-level gains, growth enabled & Connects AI to business outcomes. \\
\bottomrule
\end{tabularx}
\end{table}

\section{The Integrator Profile for Employers}
The term integrator should not be reduced to a technical job title. In the context of agentic AI, an integrator is a person who can make a system useful, safe, and accepted in the real workflow. This profile combines engineering discipline with operational empathy. It is particularly valuable for SMCs because such firms often cannot staff separate teams for AI product management, solution architecture, data governance, security, process excellence, and change management. One strong integrator can coordinate these perspectives and translate between them.

An employer can recognize integrator potential through evidence rather than slogans. A strong candidate can explain a business process before proposing a model. The candidate asks for the baseline process, exception rates, system boundaries, data ownership, user incentives, and quality criteria. The candidate can distinguish a proof of concept from a production control. The candidate is comfortable with APIs and security vocabulary but does not use technical complexity to avoid business accountability. The candidate also understands that employees are not passive recipients of automation; they are the source of process knowledge and the first line of quality assurance.

\begin{table}[!t]
\centering
\caption{Observable signals of an agentic AI integrator}
\label{tab:signals}
\scriptsize
\begin{tabularx}{\columnwidth}{P{0.28\columnwidth} P{0.33\columnwidth} Y}
\toprule
Signal & What the person does & Why employers should care \\
\midrule
Starts with process pain & Quantifies cycle time, rework, backlog, and handovers before discussing tools & Prevents AI theatre and targets measurable value. \\
Designs boundaries & Defines allowed actions, forbidden actions, thresholds, and escalation & Makes autonomy safe enough for production. \\
Understands systems & Maps ERP, CRM, DMS, e-mail, ticketing, APIs, and data sources & Converts AI from chat into workflow capability. \\
Owns verification & Builds test cases, evidence checks, sampling, and override review & Addresses the deployment verification gap. \\
Speaks governance & Uses identity, least privilege, audit, and incident language naturally & Reduces security and compliance risk. \\
Enables employees & Trains users, designs feedback loops, and communicates responsibility clearly & Builds trust and adoption quality. \\
Measures balanced value & Combines process, quality, employee, governance, and economic KPIs & Avoids narrow cost-saving logic. \\
\bottomrule
\end{tabularx}
\end{table}

For employers, the most useful interview question is not ``Which AI tools do you know?'' A better question is: ``Choose a repetitive business process, design a Level 3 agent for it, define its data sources and permissions, list the failure modes, and explain how employees remain responsible without becoming a bottleneck.'' This question reveals whether the person can integrate. It also reveals whether the person sees agentic AI as a socio-technical system rather than a model demonstration.

The same logic applies internally. Many firms already employ potential integrators: embedded engineers who understand devices and field issues, ERP key users who know process exceptions, quality managers who understand evidence and auditability, IT generalists who know interfaces and permissions, and team leads who know where employees lose time. The goal is not to centralize all AI work in a distant expert team. The goal is to develop a network of domain integrators supported by a small governance and platform core.

\subsection{Build, Buy, and Partner}
Small and medium-sized companies should be pragmatic about sourcing. They should rarely build foundation models. They may buy AI platforms, use embedded ERP or CRM agents, partner with system integrators, or develop small internal agents on approved platforms. However, the integration logic should remain internal. A vendor can provide tooling, but the firm must own process rules, data meaning, risk appetite, employee communication, and acceptance criteria. Outsourcing those decisions creates dependency and weakens learning.

A practical sourcing rule is: buy the commodity capability, own the process intelligence. Commodity capabilities include model access, orchestration platforms, document extraction, connectors, and monitoring tooling. Process intelligence includes what counts as a correct case resolution, which customer exceptions matter, when payment should be blocked, how quality deviations are interpreted, and how employees should intervene. This division protects SMCs from vendor lock-in while allowing them to move faster than a pure build strategy would allow.

\section{Discussion: What Employers Should Look For}
Employers seeking integrators should not look only for people who can demonstrate a chatbot or write impressive prompts. The limiting capability for agentic AI adoption is cross-functional integration competence: the ability to translate domain pain points into bounded, auditable, measurable human-agent workflows under real constraints. The integrator can ask uncomfortable operational questions: Which system is authoritative? What does the agent do when data conflict? Which action is reversible? Which exceptions require a human? Which data may not leave the environment? How do we know the output is correct? What will employees stop doing, and what will they start doing? What happens when the agent fails?

This skill profile is interdisciplinary. It includes enough technical competence to understand APIs, identity, retrieval, testing, logs, and system boundaries. It includes enough business competence to understand throughput, quality, service levels, and ROI. It includes enough governance competence to design controls without paralyzing the business. It includes enough change competence to bring employees into the design rather than imposing automation on them.

Small and medium-sized companies may actually be well positioned for this type of integration. Decision paths are often shorter than in large corporations, process owners are close to daily operations, and trust can be built through visible leadership. At the same time, SMCs have constraints: smaller IT teams, less formal AI governance, limited training budgets, legacy systems, and dependence on standard software vendors. That makes the integrator advantage more, not less, important. A strong integrator helps a firm avoid two traps: pilot theatre without operational value and premature autonomy without control.

\section{Outlook}
Over the next several years, agentic AI will likely move from isolated assistants toward governed portfolios of process agents. Four developments are probable. First, enterprise systems will expose more AI-ready interfaces and policy-aware connectors. Second, identity and access management will evolve to treat agents as auditable non-human actors. Third, AI literacy will become a normal part of workforce development rather than a special training event. Fourth, managerial work will shift from assigning tasks to designing human-agent systems, setting boundaries, interpreting metrics, and developing people.

For SMCs, the winning pattern is not the largest AI platform or the highest autonomy level. The winning pattern is controlled learning. Start with a painful, bounded process. Create a read-first or prepare-and-approve agent. Run shadow mode. Measure both process and employee impact. Add autonomy only when verification, monitoring, permissions, and trust are in place. Build internal integrators who can repeat the pattern across domains.

This outlook also changes career development. Employees with domain knowledge can become agent supervisors, process reviewers, knowledge curators, AI quality reviewers, or human-agent team leads. Technical employees can become agent operations owners or integration architects. Managers can become designers of learning systems. The organization that treats these roles as career paths will capture more value than the organization that treats AI as a headcount-reduction exercise.

\section{Conclusion}
Agentic AI can become a significant productivity lever for SMCs if it is positioned and integrated into bounded, measurable, and governed workflows. Its best near-term use is not full autonomy. It is controlled partial autonomy for simple and medium-complexity work that is frequent, digital, semi-structured, verifiable, and bounded in risk. Agents should handle preparation, search, classification, reconciliation, drafting, documentation, and standard-case execution. Humans should retain goal-setting, context, judgement, relationships, approvals, exception handling, quality ownership, and accountability.

The strategic capability is the integrator advantage. Employers need people who can connect business process pain with safe agent design, enterprise system integration, identity and permissions, governance, metrics, and employee enablement. In this sense, agentic AI is not merely a technology trend. It is a test of organizational integration capability.

The framework proposed in this paper is intended as a practical synthesis and should be further evaluated through empirical case studies, expert interviews, and design science validation. Such validation should test whether the scoring model, autonomy ladder, governance baseline, and upskilling stack improve use-case selection quality, deployment safety, employee acceptance, and measurable process outcomes.

The strongest position for SMCs is therefore: controlled autonomy instead of blind automation; upskilling instead of dismissal; human responsibility with agentic speed. Firms that master this balance can relieve skilled employees from repetitive work, improve quality and responsiveness, and create new development paths. Firms that treat agentic AI only as a cost-cutting instrument risk distrust, shadow AI, governance gaps, and stalled pilots. A plausible and desirable near-term future is not the peopleless company, but a human-agent organization in which skilled employees achieve more because agents are integrated, bounded, monitored, and trusted.

\balance


\begin{thebibliography}{00}
\bibitem{oecd2025sme} OECD, \emph{Generative AI and the SME Workforce: New Survey Evidence}, OECD Publishing, Paris, 2025. doi: 10.1787/2d08b99d-en. [Online]. Available: \url{https://www.oecd.org/en/publications/generative-ai-and-the-sme-workforce_2d08b99d-en.html}
\bibitem{bitkom2025} Bitkom Research, \emph{Kuenstliche Intelligenz 2025}, 2025. [Online]. Available: \url{https://bitkom-research.de/studien/kuenstliche-intelligenz-2025}
\bibitem{kfw2026ki} KfW Research, ``Einsatz von Kuenstlicher Intelligenz vor allem in Unternehmen mit hohen Innovations- und Digitalisierungsaktivitaeten verbreitet,'' \emph{Fokus Volkswirtschaft}, no. 533, Feb. 11, 2026. [Online]. Available: \url{https://www.kfw.de/PDF/Download-Center/Konzernthemen/Research/PDF-Dokumente-Fokus-Volkswirtschaft/Fokus-2026/Fokus-Nr.-533-Februar-2026-KI-Mittelstand.pdf}
\bibitem{bidt2025} bidt, ``KI im deutschen Mittelstand 2025,'' Themenmonitor, 2025. [Online]. Available: \url{https://www.bidt.digital/themenmonitor/ki-im-deutschen-mittelstand-2025/}
\bibitem{kfw2025digital} KfW Research, \emph{KfW-Digitalisierungsbericht Mittelstand 2025}, 2025. [Online]. Available: \url{https://www.kfw.de/PDF/Download-Center/Konzernthemen/Research/PDF-Dokumente-Digitalisierungsbericht-Mittelstand/KfW-Digitalisierungsbericht-2025.pdf}
\bibitem{mckinsey2025} McKinsey \& Company, \emph{The State of AI in 2025: Agents, Innovation, and Transformation}, Nov. 2025. [Online]. Available: \url{https://www.mckinsey.com/capabilities/quantumblack/our-insights/the-state-of-ai}
\bibitem{bain2025foundation} P. Gautheron, C. Bell, and S. Hardy, ``Building the foundation for agentic AI,'' Bain \& Company Technology Report, Sep. 2025. [Online]. Available: \url{https://www.bain.com/insights/building-the-foundation-for-agentic-ai-technology-report-2025/}
\bibitem{bcg2025} MIT Sloan Management Review and Boston Consulting Group, ``Managing the Machines That Manage Themselves,'' 2025. [Online]. Available: \url{https://www.bcg.com/publications/2025/machines-that-manage-themselves}
\bibitem{vu2025} H. Vu, N. Klievtsova, H. Leopold, S. Rinderle-Ma, and T. Kampik, ``Agentic business process management: Practitioner perspectives on agent governance in business processes,'' arXiv:2504.03693, 2025, preprint. [Online]. Available: \url{https://arxiv.org/abs/2504.03693}
\bibitem{apostolou2026} S. A. Apostolou, J. Bosch, and H. Holmstroem Olsson, ``Agentic AI in industry: Adoption level and deployment barriers,'' arXiv:2605.14675, 2026, preprint. [Online]. Available: \url{https://arxiv.org/abs/2605.14675}
\bibitem{azarijafari2025} M. Azarijafari, L. Mich, and M. Missikoff, ``An agentic AI for a new paradigm in business process development,'' arXiv:2507.21823, 2025, preprint. [Online]. Available: \url{https://arxiv.org/abs/2507.21823}
\bibitem{ilo2025} International Labour Organization, \emph{Generative AI and Jobs: A 2025 Update}, 2025. [Online]. Available: \url{https://www.ilo.org/publications/generative-ai-and-jobs-2025-update}
\bibitem{wef2025} World Economic Forum, \emph{The Future of Jobs Report 2025}, Jan. 2025. [Online]. Available: \url{https://www.weforum.org/publications/the-future-of-jobs-report-2025/}
\bibitem{pwc2026} PwC, \emph{2026 AI Global Jobs Barometer: Two Futures for Jobs in an AI Era}, Jun. 2026. [Online]. Available: \url{https://www.pwc.com/gx/en/services/ai/ai-jobs-barometer.html}
\bibitem{microsoft2026} Microsoft, \emph{2026 Work Trend Index: Agents, Human Agency, and the Opportunity for Every Organization}, May 2026. [Online]. Available: \url{https://www.microsoft.com/en-us/worklab/work-trend-index/agents-human-agency-and-the-opportunity-for-every-organization}
\bibitem{euaiart4} European Commission, AI Act Service Desk, ``Article 4: AI literacy,'' 2025. [Online]. Available: \url{https://ai-act-service-desk.ec.europa.eu/en/ai-act/article-4}
\bibitem{eugpai2025} European Commission, ``General-purpose AI obligations under the AI Act,'' 2025. [Online]. Available: \url{https://digital-strategy.ec.europa.eu/en/factpages/general-purpose-ai-obligations-under-ai-act}
\bibitem{nist2024} NIST, \emph{Artificial Intelligence Risk Management Framework: Generative Artificial Intelligence Profile}, NIST AI 600-1, Jul. 2024. [Online]. Available: \url{https://www.nist.gov/publications/artificial-intelligence-risk-management-framework-generative-artificial-intelligence}
\bibitem{iso42001} ISO, ``ISO/IEC 42001:2023 - Artificial intelligence management systems,'' 2023. [Online]. Available: \url{https://www.iso.org/standard/42001}
\bibitem{owasp2025} OWASP GenAI Security Project, ``OWASP Top 10 for Agentic Applications,'' 2025. [Online]. Available: \url{https://genai.owasp.org/2025/12/09/owasp-top-10-for-agentic-applications-the-benchmark-for-agentic-security-in-the-age-of-autonomous-ai/}
\bibitem{openid2025} OpenID Foundation, \emph{Identity Management for Agentic AI}, 2025. [Online]. Available: \url{https://openid.net/wp-content/uploads/2025/10/Identity-Management-for-Agentic-AI.pdf}
\bibitem{csa2026} Cloud Security Alliance, ``More than two-thirds of organizations cannot clearly distinguish AI agent from human actions,'' Press release, Mar. 24, 2026. [Online]. Available: \url{https://cloudsecurityalliance.org/press-releases/2026/03/24/more-than-two-thirds-of-organizations-cannot-clearly-distinguish-ai-agent-from-human-actions}
\end{thebibliography}
\end{document}